\documentclass[runningheads]{llncs}
\usepackage[T1]{fontenc}
\usepackage{graphicx}
\usepackage{xspace}
\newcommand{\ODD}{\textit{O\kern-0.1em D\kern-0.1em D}\xspace}
\DeclareRobustCommand{\method}[1]{{\fontsize{8}{12}\selectfont \textbf{#1}}}
\usepackage{amsmath}
\usepackage{amssymb}

\newcommand{\FeatF}{{\scriptstyle \mathbf{F}}}
\newcommand{\FeatL}{{\scriptstyle \mathbf{L}}}
\newcommand{\FeatR}{{\scriptstyle \mathbf{R}}}

\usepackage{xcolor}
\usepackage{amsmath,amssymb}
\usepackage{algorithm}
\usepackage{algpseudocode}

\usepackage{url}
\usepackage{hyperref}

\usepackage{wrapfig}

\usepackage{subcaption} 

\usepackage{booktabs}   
\usepackage{array}      
\usepackage{caption}    
\usepackage{multirow}
\usepackage{adjustbox}

\begin{document}
\title{Online Domain-aware LLM Decoding for Continual Domain Evolution}
\author{Mohammad Abu-Shaira \and Weishi Shi}
\authorrunning{M. Abu-Shaira and W. Shi}
\titlerunning{Online Domain-aware LLM Decoding}
\institute{
University of North Texas, Denton, TX, USA
}
%
%
%
\maketitle              
\begin{abstract}
LLMs are typically fine-tuned offline on domain-specific data, assuming a static domain. In practice, domain knowledge evolves continuously through new regulations, products, services, and interaction patterns. Retraining or fine-tuning LLMs for every new instance is computationally infeasible. Additionally, real-world environments also exhibit temporal dynamics with shifting data distributions. Disregarding this phenomenon, commonly referred to as concept drift, can significantly diminish a model’s predictive accuracy. This mismatch between evolving domains and static adaptation pipelines highlights the need for efficient, real-time adaptation without costly retraining. In response, we introduce Online Domain-aware Decoding framework (\ODD). \ODD performs probability-level fusion between a base LLM and a prefix-tree prior, guided by adaptive confidence modulation using disagreement and continuity signals.
Empirical evaluation under diverse drift scenarios demonstrates that \ODD consistently surpasses LLM-Greedy and LLM-Temp Scaled across all syntactic and semantic NLG metrics. It yields an absolute ROUGE-L gain of $\mathbf{0.065}$ and a $\mathbf{13.6\%}$ relative improvement in Cosine Similarity over the best baseline. These results demonstrate \ODD’s robustness to evolving lexical and contextual patterns, making it suitable for dynamic LLM applications.

\end{abstract}

\section{Introduction}
\vspace{-2pt}
Large Language Models (LLMs) such as GPT-3 \cite{brown2020language}, PaLM \cite{chowdhery2023palm}, and LLaMA \cite{touvron2023llama} excel in many NLP tasks due to large-scale pretraining and adaptation techniques like supervised fine-tuning and RLHF \cite{ouyang2022training}. While effective for generalization, these offline strategies are limited when deployed in specialized domains that require strict adherence to dynamic, evolving knowledge. For example, LLM-powered assistants in customer support and decision-making \cite{zhang2020dialogpt} demand consistency, factual accuracy, and domain compliance. Yet both general-purpose and fine-tuned models often hallucinate or deviate from requirements when domain knowledge changes \cite{ji2023survey,maynez-etal-2020-faithfulness}.

Although pretrained LLMs are typically fine-tuned on \textit{domain-specific} datasets, this adaptation is performed \textit{offline} and assumes a static domain. In practice, domain knowledge evolves through new terminology, updated products, changing regulations, shifting pricing, modified services, and emerging user intents, causing \textit{concept drift} \cite{gama2014survey}. Because modern LLMs are expensive to retrain, frequent fine-tuning is impractical, creating a barrier to real-time adaptation \cite{shi2024continual}. This gap underscores the need for efficient, real-time learning approaches \cite{gama2014survey,de2021continual}. Despite their versatility, LLMs face practical challenges in dynamic environments. Repeated fine-tuning or retraining is computationally expensive and time-consuming \cite{zhao2023survey}, risks \textit{catastrophic forgetting} \cite{de2021continual}, and may violate \textit{data privacy and regulatory} constraints. These limitations motivate inference-time mechanisms that incorporate new domain knowledge without modifying model weights.

Although extensive research exists, current decoding and adaptation strategies exhibit fundamental limitations for online domain adaptation. \textit{Logit-based decoding control} avoids retraining but often requires auxiliary controllers or extra forward/backward passes per token, providing only local control that degrades under domain shift. \textit{Prefix- and context-aware decoding} methods impose rigid constraints or rely on static priors, lacking mechanisms for temporal or drift-aware adaptation. \textit{Retrieval-guided decoding} introduces significant latency overhead and relies on the freshness and maintenance of external knowledge indices. Finally, \textit{parameter-efficient adaptation} and model-editing techniques require offline, task-specific training or permanent modification of model weights, preventing real-time, zero-training inference adaptation.

These limitations highlight the need for a drift-aware framework that enables online adaptability without retraining, auxiliary models, or external retrieval. In response, we introduce \ODD{}, a lightweight inference-time framework that dynamically aligns generation with evolving domain knowledge. \ODD{} performs probability-level fusion between the base LLM and an online Trie prior, guided by adaptive \textit{confidence} modulation through \textit{disagreement} and temporal \textit{continuity} signals. The source code is available at our GitHub repository \cite{ODD-GIT}.

\section{Related Work}
\vspace{-2pt}
\textbf{Logit-Based Decoding Control: } Decoding control adjusts logits during inference to steer generation. \emph{Auxiliary controllers} (e.g., \textit{PPLM}~\cite{dathathriplug}, \textit{GeDi}~\cite{Krause2021GeDi}) rely on external models, while \emph{internal methods} (\textit{DExperts}~\cite{Liu2021DExperts}, \textit{Self-Debiasing}~\cite{schick2021self}, \textit{FUDGE}~\cite{Yang2021FUDGE}) use contrastive or self-generated signals. These techniques avoid full retraining but demand auxiliary passes and offer only short-term control that fails under domain shift.
\textbf{Prefix and Context-Aware Decoding: } Prefix-aware decoding guides next-token prediction using the generated context. \emph{Hard constraints} (e.g., \textit{Constrained Beam Search}~\cite{anderson2017guided}) enforce required phrases, while \emph{soft priors} (e.g., \textit{Dynamic Vocabulary Selection/Pruning}~\cite{l2016vocabulary,wu2018neural}) reweight token probabilities based on prefix-conditioned candidates.
\textbf{Retrieval- and Knowledge Guided Decoding}
These methods retrieve external evidence conditioned on the prefix and integrate it to bias next-token prediction. \textit{kNN-LM}~\cite{khandelwalgeneralization} interpolates the base distribution with nearest-neighbor targets, while generative models such as \textit{RAG}~\cite{lewis2020retrieval} and \textit{RETRO}~\cite{borgeaud2022improving} condition decoding on retrieved documents for fact-aware output. \textit{Knowledge-Infused Decoding}~\cite{liuknowledge} instead injects knowledge signals directly into the logits. These methods provide strong factual grounding but incur retrieval latency and depend on index freshness.
\textbf{Parameter-Efficient Adaptation and Model Editing}
\emph{Parameter-Efficient Adaptation (PEA)} trains lightweight components while freezing the base LLM. Methods such as \textit{Prefix-Tuning}~\cite{li2021prefix}, \textit{Prompt Tuning}~\cite{lester2021power}, and \textit{LoRA}~\cite{hulora} adapt internal representations through task-specific training. In contrast, \emph{Model Editing} techniques like \textit{ROME}~\cite{meng2022locating} and \textit{MEMIT}~\cite{meng2023mass} modify model parameters directly to embed factual updates. Both lines of work complement decoding-based approaches that guide inference externally without altering model weights.

\section{Problem Settings}
\vspace{-2pt}
Let $\mathcal{V}$ be the vocabulary and $\mathbf{x} = (x_1, \dots, x_T)$ be a sequence. Given a prefix $\pi_t = (x_1, \dots, x_t)$, the LLM (parameterized by $\theta$) estimates the next token distribution $P_\theta(y \mid \pi_t)$ from its output logit vector $\mathbf{z}_t \in \mathbb{R}^{|\mathcal{V}|}$. We define the base distribution as $q_t^{\mathrm{LM}} = \mathrm{softmax}(\mathbf{z}_t)$, where $q_t^{\mathrm{LM}} = P_\theta(\cdot \mid \pi_t)$. 
LLMs are typically fine-tuned offline assuming a stationary data distribution. In real-world scenarios, domain knowledge evolves continuously, with new sequences arriving sequentially in an unbounded stream. This non-stationary distribution leads to \textit{concept drift}. Formally, drift occurs when $\exists \, \pi_t, y : \; p_{t}(y \mid \pi_t) \neq p_{t+1}(y \mid \pi_t),$ meaning the true distribution over next-token continuations changes over time. This challenge requires maintaining LLM performance in online environments. 
This necessitates adapting the LLM's next-token distribution in real time to evolving domain knowledge, critically \textit{avoiding parameter updates}. Our objective is to construct an adapted distribution $\tilde{q}_t$ that balances fidelity to the base $q_t^{\mathrm{LM}}$ (\textit{stability}) with responsiveness to domain evidence from the \textit{Prefix Trie} (\textit{adaptability}). Formally, the objective balances \textit{stability} to the base model with \textit{adaptability} to domain-specific signals extracted from the Trie. Let $q_t^{\mathrm{Pn}}$ denote the Trie-induced prior derived from the feature vector $\phi_t$. We express this trade-off as:
\vspace{-5pt}
\[
\tilde{q}_t
= \arg\min_{q \in \Delta^{|\mathcal{V}|}}
\Big\{
\mathcal{D}(q \,\|\, q_t^{\mathrm{LM}})
+
\mathcal{D}(q \,\|\, q_t^{\mathrm{Pn}})
\Big\},
\]
where $\mathcal{D}(\cdot\|\cdot)$ is a divergence measure capturing stability, $q_t^{\mathrm{Pn}}$ encodes the domain-specific Trie features through $\phi_t$, and the formulation jointly encourages adherence to the base model while aligning with evolving domain evidence.


\section{Method}
\vspace{-2pt}
The proposed decoding strategy augments the finetuned base LLM with domain priors extracted from the Prefix Trie. At each generation step, the strategy constructs two distributions: the LLM's base distribution and the Prefix Trie prior, which captures structural statistics (\textit{frequency, recency, and length}) from the online data stream. These distributions are combined through an adaptive fusion mechanism using a \textit{confidence-based} weighting factor, which is dynamically determined by \textit{disagreement} and temporal \textit{continuity} signals. This fusion leverages the LLM’s finetuned domain knowledge while incorporating the Trie’s fast, online updates, circumventing the high cost of continual finetuning. To establish a strong domain foundation, we \textit{finetune a pretrained LLM} on target data. This model serves as the base for our approach, supplying the output logits ($\mathbf{z}_t \in \mathbb{R}^{|V|}$) that initiate the fusion process at every prediction step.


\subsection{Online Prefix Tree Construction}
\vspace{-2pt}
The Prefix Tree (Trie) is continuously updated with evolving domain knowledge using an $n$-gram scheme that inserts all token sequences up to length $N$. This keeps both short and long domain phrases current and directly available for next-token prediction.

\begin{wrapfigure}{r}{0.53\columnwidth} 
\vspace{-30pt} 
\centering
\scriptsize
\setlength{\baselineskip}{0.85\baselineskip}
\begin{verbatim}
(root)
  +-- activate (F=2, L=1, R=1759263260)
  |   +-- your (F=2, L=2, R=1759263260)
  |       +-- plan (F=2, L=3, R=1759263260)
  |           |__ 4G (F=1, L=4, R=1758658460)
  |           |__ 5G (F=1, L=4, R=1759263260)
  +-- your (F=2, L=1, R=1759263260)
  |   +-- plan (F=2, L=2, R=1759263260)
  |       |__ 4G (F=1, L=3, R=1758658460)
  |       |__ 5G (F=1, L=3, R=1759263260)
  +-- plan (F=2, L=1, R=1759263260)
  |   |__ 4G (F=1, L=2, R=1758658460)
  |   |__ 5G (F=1, L=2, R=1759263260)
  +-- please (F=1, L=1, R=1759263260)
      +-- activate (F=1, L=2, R=1759263260)
          +-- your (F=1, L=3, R=1759263260)
              +-- plan (F=1, L=4, R=1759263260)
                  |__ 5G (F=1, L=5, R=1759263260)
\end{verbatim}
\vspace{-16pt}
\caption{\centering\scriptsize $N$-gram Trie with Frequency ($\FeatF$), Length ($\FeatL$), and Recency ($\FeatR$).}
\label{fig:prefix-tree}
\vspace{-15pt}
\end{wrapfigure}
Each trie entry stores three features: $\textit{Recency}$ ($\FeatR$), $\textit{Length}$ ($\FeatL$), and $\textit{Frequency}$ ($\FeatF$). $\FeatR$ tracks when a continuation last appeared, $\FeatL$ measures how closely its span matches the query prefix, and $\FeatF$ counts how often it occurs. These features are updated at each node whenever new instances appear, allowing the trie to function as a dynamic scoring mechanism that reflects the temporal, structural, and statistical relevance of domain knowledge.
Consider two domain sequences inserted into the $n$-gram Trie: the older $\mathbf{s_1}$ (``Activate your plan 4G'') and the more recent $\mathbf{s_2}$ (``Please activate your plan 5G''). \textit{For clarity, this illustration uses word-level nodes, while the actual implementation constructs the Trie over token-level sequences}. The node-level features ($\FeatF, \FeatL, \FeatR$) capture the structural, statistical, and temporal properties of these sequences (see Figure~\ref{fig:prefix-tree}): \textbf{Recency ($\FeatR$):} Records the most recent observation time. The ``4G'' leaf carries an earlier timestamp, whereas ``5G'' reflects the newer update, enabling recency-aware scoring.
\textbf{Length ($\FeatL$):} Denotes the depth of the token in the sequence, rewarding longer, more specific matches (e.g., $L=5$ for the full ``5G'' path).
\textbf{Frequency ($\FeatF$):} Counts the number of times a prefix appears. For example, the shared path ``activate $\rightarrow$ your $\rightarrow$ plan'' has $\FeatF = 2$.

\subsection{Candidate Tokens Scoring Mechanism}
\vspace{-2pt}
At each decoding step $t$, the method queries the trie to construct a candidate set $\mathcal{C}$ (unique next tokens with highest feature score), and computes a probability distribution $q_t^{\mathcal{P}_n}$ over $\mathcal{C}$. Given the prefix $\pi_t$, \textit{all matching suffixes} in the Trie $\mathcal{P}_n$ are queried to populate the candidate set $\mathcal{C}$ with unique next tokens $y$ and their associated feature vectors $\phi = (\FeatF, \FeatL, \FeatR)$. To reduce bias from large token counts, the \textbf{Frequency} feature is transformed as $\FeatF''=\log(1+\FeatF)$. Each candidate’s preliminary score is computed from the normalized feature tuple $(\FeatF', \FeatL', \FeatR')$. Frequency is normalized across the candidate set using $\FeatF'=\FeatF'' / \max_{y' \in \mathcal{C}} \FeatF''$, ensuring that frequent continuations do not dominate the scoring range and remain comparable to the other normalized features. \textbf{Length} is normalized with respect to the current prefix depth, $\FeatL' = \FeatL / |\pi_t|$, ensuring proportional credit for contextual alignment rather than rewarding overly long sequences. \textbf{Recency} is modeled via an exponential decay, $\FeatR'=\exp(-\Delta/\Delta_{\max})$, where $\Delta$ denotes the recency gap; here, $\Delta_{\max} = \max_{y' \in \mathcal{C}} \Delta(y')$ ensures well-defined normalization over the candidate set. This assigns $\FeatR'=1$ to the most recent continuation while preserving a nonzero contribution for older candidates. The final score for each candidate token is obtained through a weighted linear combination, $\mathbf{score}(y)=\langle \boldsymbol{\lambda}, (\FeatF', \FeatL', \FeatR') \rangle.$ The non-negative weight vector $\boldsymbol{\lambda} = (\lambda_R, \lambda_L, \lambda_F)$ (summing to $1$) is configurable, allowing users to prioritize specific domain statistics (e.g., \textit{Recency}). This formulation yields an interpretable final score in the range $(0,1]$, reflecting the candidate's overall strength. The resulting scores are later converted into a valid probability distribution $q_t^{\mathcal{P}_n}$ by top-preserving normalization over the candidate set $\mathcal{C}$, assigning zero mass to tokens outside $\mathcal{C}$.

\subsection{Probability Distribution Preparation and Calibration}
\vspace{-2pt}
The candidate scores are first mapped to the sparse Trie distribution ($q_t^{\mathcal{P}_n}$) using a \textit{top-preserving normalization} scheme (Line 17, Algorithm \ref{alg:spine}), which retains the maximum score ($S_{\max}$) while proportionally distributing the remaining mass to ensure a valid probability sum. Since the neural $q_t^{\text{LM}}$ and statistical $q_t^{\mathcal{P}_n}$ distributions operate on different confidence scales, we apply \textit{adaptive temperature scaling} prior to interpolation. This mechanism dynamically adjusts the LLM's sharpness (via $T_t^{*}$ in Equation \ref{eq:T}) until its peak probability precisely matches that of $q_t^{\mathcal{P}_n}$ (Equation \ref{eq:equal_dist_maxes}). This crucial calibration eliminates scale bias between the experts, preparing them for the final dynamic fusion ($\tilde{q}_t = \gamma_t q_t^{\text{LM}} + (1-\gamma_t) q_t^{\mathcal{P}_n}$). Because temperature scaling is a monotonic, rank-preserving transformation of logits, it adjusts only the sharpness of the distribution and therefore modifies probability mass in a controlled manner.
{\footnotesize
\begin{equation}
\label{eq:T}
    T_t^{*} =
    \arg\min_{T > 0}
    \Bigg|
    \max_{y \in V}\; \mathrm{softmax}\!\Big(\tfrac{\mathbf{z}_t}{T}\Big)
    -
    \max_{y \in V}\; q_t^{\mathcal{P}_n}(y)
    \Bigg|
\end{equation}\vspace{-8pt}
\begin{equation}
\label{eq:equal_dist_maxes}
    \max_{y \in V}\; \mathrm{softmax}\!\Big(\tfrac{\mathbf{z}_t}{T_t}\Big)
    =
    \max_{y \in V}\; q_t^{\mathcal{P}_n}(y)
\end{equation}
}
A unique solution for $T_t$ always exists because $\max \mathrm{softmax}(\mathbf{z}_t / T)$ is a continuous, strictly monotonic function of $T$, and we obtain $T_t$ via a 1D monotonic root-finding method (bisection).
\begin{algorithm}[t!]
\caption{\ODD : Online Domain-Aware Decoding}
\label{alg:spine}
\footnotesize
\setlength{\intextsep}{0pt}
\setlength{\textfloatsep}{0pt}
\begin{algorithmic}[1]

\State \textbf{Input:} $\mathbf{z}_t \in \mathbb{R}^{|V|};\,
\mathcal{P}_n = \{\, p \mapsto n(p) : \phi \mid \phi = (\FeatF,\FeatL,\FeatR)\,\};\,
\pi_t = (x_1,\dots,x_t);\,
\theta = \{\boldsymbol{\lambda}\}$
\hspace*{1em}\Comment{\footnotesize \parbox[t]{.45\linewidth}{\textcolor{blue}{%
$\boldsymbol{\lambda} = (\lambda_{\FeatF},\lambda_{\FeatL},\lambda_{\FeatR}),\; \sum_i \lambda_i = 1$;\;
$x_i \in V$}}}

\State $\mathcal{C} \gets \{\;\}$ 
\Comment{\footnotesize \parbox[t]{.45\linewidth}{\textcolor{blue}{$\mathcal{C}$ is the candidate set}}}

\For{each suffix $s$ of $\pi_t$}
    \For{each $(y,\phi) \in \method{NextTokens}(s, \mathcal{P}_n)$}
    


        \State $\FeatF'' = \log(1+\FeatF)$;
        \quad\quad $\FeatF' \gets \dfrac{\FeatF''}{\max\limits_{y' \in \mathcal{C}} \FeatF''}$; \quad\quad $\FeatL' \gets \dfrac{\FeatL}{|\pi_t|}$;
        \quad\quad $\FeatR' \gets \exp\!\Big(-\dfrac{\Delta}{\Delta_{\max}}\Big)$

        \State $\mathrm{score}(y) \gets 
        \langle \boldsymbol{\lambda}, (\FeatF', \FeatL', \FeatR') \rangle$\Comment{\footnotesize \parbox[t]{.30\linewidth}{\textcolor{blue}{$\mathrm{score}(y) \in (0,1]$}}}

        \If{$y \notin \mathcal{C}$ \textbf{or} $\mathrm{score}(y) > \mathrm{score}_{\mathcal{C}}(y)$}
            \State update $\mathcal{C}$ with $(y,\mathrm{score}(y))$
        \EndIf
    \EndFor
\EndFor


\State $q_t^{\text{LM-base}} \gets \mathrm{softmax}\!\big(\mathbf{z}_t\big)$\Comment{\footnotesize \parbox[t]{.60\linewidth}{\textcolor{blue}{%
LLM distribution (before temp adjustment)}}}


\State $H \gets -\sum_{y \in V} q_t^{\text{LM-base}}[y] \log q_t^{\text{LM-base}}[y]$

\State $c_{\text{LM}} \gets 1 - \tfrac{H}{H_{\max}}$\Comment{\footnotesize \parbox[t]{.30\linewidth}{\textcolor{blue}{%
$H_{\max} \gets \log |V|$}}}


\State $q_t^{\mathcal{P}_n}(y) \gets 
S_{\max}\,\mathbf{1}_{\{y = y_{\max}\}} 
+ \dfrac{(1 - S_{\max})\,\mathrm{score}(y)}
{\sum_{y' \in \mathcal{C} \setminus \{y_{\max}\}} \mathrm{score}(y')}$ 
\Comment{\footnotesize \parbox[t]{.30\linewidth}{\textcolor{blue}{$S_{\max} = \displaystyle\max_{y \in \mathcal{C}} \mathrm{score}(y)$}}}

\State $c_{\text{trie}} \gets \max_{y \in V} q_t^{\mathcal{P}_n}[y]$

\State $q_t^{\text{LM}} \gets \mathrm{softmax}\!\big(\tfrac{\mathbf{z}_t}{T_t}\big)$ 
\Comment{\footnotesize \parbox[t]{.65\linewidth}{\textcolor{blue}{%
LLM distribution (after adaptive temperature). $T_t$ is adaptive temperature chosen s.t.\ $\max q_t^{\text{LM}} = S_{\max}$}}}


\State $c'_{\text{LM}} \gets c_{\text{LM}} (1 - \Omega^2)$\Comment{\footnotesize \parbox[t]{.65\linewidth}{\textcolor{blue}{$\Omega$: disagreement using Top-$k$ (here $k=5$)}}}

\State $c'_{\text{trie}} \gets c_{\text{trie}} + (1 - c_{\text{trie}})\,c_{\text{trie}}^{2} \, \Gamma$\Comment{\footnotesize \parbox[t]{.35\linewidth}{\textcolor{blue}{$\Gamma$ is continuity}}}

\State $\gamma_t \gets \dfrac{c'_{\text{LM}}}{c'_{\text{LM}} + c'_{\text{trie}}}$

\State $y^{*} \gets \arg\max_{y \in V} \;\Big(\gamma_t\,q_t^{\text{LM}}[y] + (1-\gamma_t)\,q_t^{\mathcal{P}_n}[y]\Big)$

\State $\pi_{t+1} \gets (\pi_t, y^*)$
\State \textbf{return} $\pi_{t+1}$

\end{algorithmic}
\end{algorithm}
\subsection{Establishing Distributions Confidence}
\vspace{-2pt}
The adaptive interpolation weight $\gamma_t$ is computed dynamically at each step, reflecting the current certainty of the two experts. When the LLM is confident, $\gamma_t$ is high, increasing its influence, and vice versa. The base \textit{LLM confidence} ($c_{\text{LM}}$) is quantified using its \textit{normalized entropy} ($H(\mathbf{q}_t^{\text{LM}})$) (Lines 15, and 16 Algorithm \ref{alg:spine}), resulting in a scale-invariant measure $c_{\text{LM}} \in [0,1]$ where low entropy (high certainty) results in $c_{\text{LM}}$ near 1. The \textit{Trie confidence} ($c_{\text{trie}}$) is simply defined as the \textit{maximum probability} assigned to its top-preserving candidate (Line 18, Algorithm \ref{alg:spine}). Since $q_t^{\mathcal{P}_n}$ is constructed to emphasize its strongest candidate, this maximum value naturally indicates the distribution's confidence.

\subsection{Rewarding and Penalizing Distribution Confidence}
\vspace{-2pt}
The intrinsic confidences ($c_{\text{LM}}$ and $c_{\text{trie}}$) are refined using two context-aware factors: \textbf{Disagreement ($\Omega_t$)} and \textbf{Continuity ($\Gamma_t$)}. $\Omega_t$ measures how differently the two experts rank their top-$k$ tokens (we use $k=5$ in all experiments), while $\Gamma_t$ captures temporal agreement through $r_t$, the number of consecutive steps in which both experts select the same top token. We form the union set $\mathcal{I}_k$ from the top-$k$ tokens of each distribution and normalize them over this set (Equations \ref{eq:normalized_pq1}--\ref{eq:normalized_pq2}). JSD is computed (Equation \ref{eq:jsd}), and $\Omega_t$ follows from its square root (Equation \ref{eq:D}), with larger values reducing the LLM's confidence. $\Gamma_t$ is an exponential function of $r_t$ (Equation \ref{eq:C}), rewarding the Trie when agreement is consistent. The final confidences ($c'_{\text{LM}}$, $c'_{\text{trie}}$) combine these factors, with $\Omega_t$ penalizing the LLM and $\Gamma_t$ amplifying the Trie.

{\footnotesize
\setlength{\abovedisplayskip}{0pt}
\setlength{\belowdisplayskip}{0pt}
\setlength{\abovedisplayshortskip}{0pt}
\setlength{\belowdisplayshortskip}{0pt}
\noindent
\begin{minipage}{0.48\linewidth}
\begin{equation}
\label{eq:normalized_pq1}
p(y) = \frac{q_t^{\text{LM}}[y]}{\sum_{y' \in \mathcal{I}_k} q_t^{\text{LM}}[y']}
\end{equation}
\end{minipage}
\hfill
\begin{minipage}{0.48\linewidth}
\begin{equation}
\label{eq:normalized_pq2}   
q(y) = \frac{q_t^{\mathcal{P}_n}[y]}{\sum_{y' \in \mathcal{I}_k} q_t^{\mathcal{P}_n}[y']}
\end{equation}
\end{minipage}
\vspace{1pt}
\noindent
\begin{equation}
\label{eq:jsd}
\mathrm{JSD}(p,q)
= \tfrac{1}{2} \sum_{y \in \mathcal{I}_k}
\big[
p(y) \log \tfrac{p(y)}{m(y)} +
q(y) \log \tfrac{q(y)}{m(y)}
\big],
\quad m(y) = \tfrac{1}{2}\big(p(y) + q(y)\big)
\end{equation}
\noindent
\begin{minipage}{0.48\linewidth}
\begin{equation}
\label{eq:D}
\Omega_t = \min\!\big(1, \sqrt{\mathrm{JSD}(p,q)}\big)
\end{equation}
\end{minipage}
\hfill
\begin{minipage}{0.48\linewidth}
\begin{equation}
\label{eq:C}
\Gamma_t = 1 - \exp\!\Big(-\frac{r_t}{3}\Big)
\end{equation}
\vspace{-10pt}
\end{minipage}
}

\subsection{Next Token Prediction}
\vspace{-2pt}
Finally, the adaptive interpolation coefficient ($\gamma_t$) is computed from the adjusted confidences, preserving \textbf{relative confidence} by ensuring $\gamma_t = c'_{\text{LM}} / (c'_{\text{LM}} + c'_{\text{trie}})$. This coefficient is used to form the final adapted probability distribution ($\tilde{q}_t$) as a convex combination of the LLM's linguistic prior and the Trie's likelihoods: $\tilde{q}_t(y) = \gamma_t \, q_t^{\text{LM}}(y) + (1 - \gamma_t) \, q_t^{\mathcal{P}_n}(y)$. At each step, the next token $y_t^{*}$ is selected by maximizing this mixed distribution, $y_t^{*} = \arg\max_{y \in V} \, \tilde{q}_t(y)$, guaranteeing a principled balance between linguistic fluency and domain-specific recurrence.

\subsection{Time and Memory Complexity}
\vspace{-2pt}
The operational efficiency of the Trie is fundamental to \ODD's lightweight, inference-time design. Defining $L = |\pi_t|$ as the prefix length, \textbf{online insertion} remains linear, $\mathcal{O}(L)$, as each token corresponds to one node traversal. The \textbf{all-suffix retrieval} procedure explores every suffix of the prefix sequence $\pi_t$, resulting in a worst-case complexity of $\mathcal{O}(L^2)$. Importantly, both insertion and retrieval are \textbf{independent of the number of stored sequences ($N$)}, ensures decoding latency remains bounded as the domain knowledge base expands. 

Memory-wise, the Trie allocates one node per distinct token position. The worst-case memory complexity is $\mathcal{O}(L_{\mathrm{total}})$, where $L_{\mathrm{total}}$ is the total number of inserted tokens. However, because \ODD indexes only compact, tokenized placeholders rather than full conversational texts, the trie's size grows proportionally to the number of unique placeholder patterns. The resulting \textbf{empirical memory complexity is $\mathcal{O}(U)$}, where $U$ denotes the number of distinct placeholder prefixes, which is typically far smaller than the theoretical bound. Empirical profiling confirms that Trie updates and retrievals introduce sub-millisecond latency per decoding step, consistent with the theoretical bounds above.

\section{Experiments}
\vspace{-2pt}
\subsection{Datasets, Evaluation Metric \& Baseline}
\vspace{-2pt}
We use the \textit{Bitext Telco LLM Chatbot Training Dataset} ($\approx 26{,}000$ samples, 26 intents), which \textbf{natively contains placeholder tokens}. We substitute these placeholders to create controlled \textit{abrupt, incremental, and gradual} concept-drift scenarios. Figure~\cite{bitext_telco_llm} illustrates the \textit{Bitext} dataset under the \textit{abrupt, incremental, and gradual} placeholder drift scenario. To measure the overall distributional drift, we track lexical and semantic shifts using \textit{Jensen--Shannon Distance (JSD)} to assess lexical divergence, \textit{Cosine Distance} to capture semantic drift on sentence embeddings, \textit{BERTScore F1} to assess contextual alignment, and \textit{Maximum Mean Discrepancy (MMD)}. To evaluate \textit{model performance} under drift, we compare \ODD against baseline decoders using: (a) lexical similarity (Exact Match\cite{rajpurkar2016squad}, NED~\cite{yujian2007normalized}, BLEU~\cite{papineni2002bleu}, ChrF~\cite{popovic2015chrf}), (b) structural coherence (ROUGE-L~\cite{lin2004rouge}), and (c) semantic alignment (Cosine Similarity~\cite{reimers2019sentence}, BERTScore~\cite{zhang2020bertscore}). The dataset involves placeholder tokens, a common convention in public and real-world NLP datasets such as MultiWOZ and customer-support corpora, used to represent dynamic and evolving entities (for example, product names or account identifiers). This placeholder drift provides a practical and realistic proxy for real-world distribution shift. Importantly, \ODD is not limited to placeholders; the Trie incorporates any newly observed tokens or patterns, enabling domain-agnostic adaptation. Both calibration and confidence computation add less than 0.2 ms overhead per decoding step, preserving real-time inference.

We benchmark our method against two well-recognized decoding strategies: \textit{LLM-Greedy} and \textit{LLM-Temp Scaled}. We restrict our baselines to decoding-only methods because our setting assumes zero-training, inference-time online adaptation with no external retrieval or model updates. Retrieval-augmented approaches (e.g., RAG, kNN-LM), prefix-constrained decoding (e.g., Constrained Beam Search, CBS), and parameter-efficient adaptation methods require external memory, static constraints, or offline training, making them incompatible with our real-time drift scenario.

\begin{figure}[t]
    \centering
    \includegraphics[width=.85\linewidth, height=0.25\textheight]{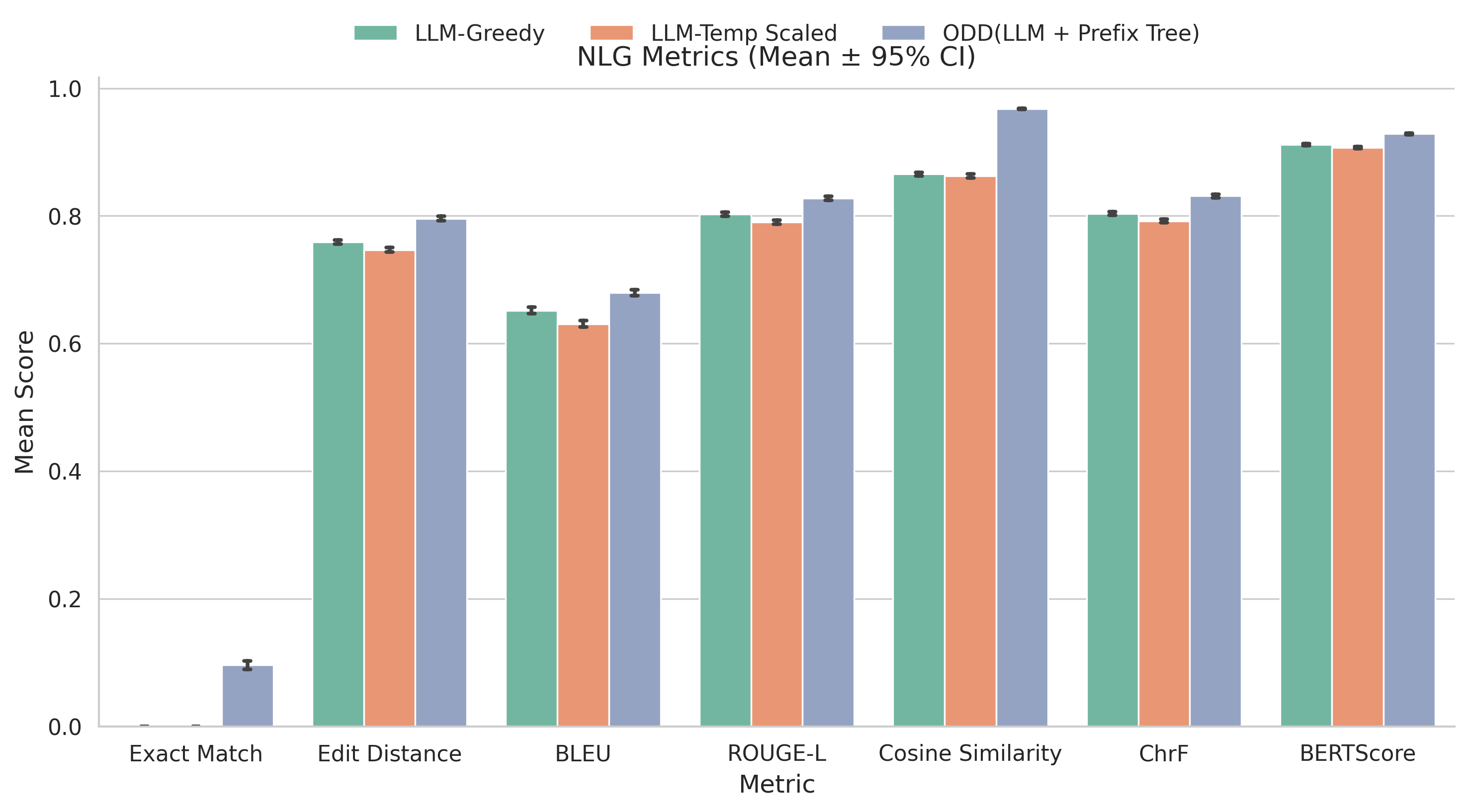}
    \vspace{-15pt}\caption{\centering\scriptsize NLG Metrics (Mean $\pm$ 95\% CI) under abrupt drift}
    \label{fig:nlg_barplot_abrupt}
    \vspace{-12pt}
\end{figure}
\begin{table}[htbp]
\vspace{-5mm} 
\centering
\caption{\centering\textbf{Decoding Strategies Performance Under Drift}\\
{\scriptsize E.M.\ Exact Match,
E.D.\ Edit Distance, Cos-Sim.\ Cosine Similarity, BERT.\ BERTScore.}}
\label{tab:drift_results}
\renewcommand{\arraystretch}{.9}
\begin{adjustbox}{width=\linewidth}
\begin{tabular}{lccccccc}
\toprule
{\scriptsize \textbf{Strategy}} &
{\scriptsize \textbf{E.M.}} &
{\scriptsize \textbf{E.D.}} &
{\scriptsize \textbf{BLEU}} &
{\scriptsize \textbf{ROUGE-L}} &
{\scriptsize \textbf{Cosine Sim.}} &
{\scriptsize \textbf{ChrF}} &
{\scriptsize \textbf{BERT.}} \\
\midrule
\multicolumn{8}{c}{\textbf{\footnotesize Abrupt Drift}} \\
\textbf{Greedy}        & 0     & 0.759 & 0.652 & 0.803 & 0.865 & 80.374 & 0.911 \\
\textbf{Temp Scaled}   & 0     & 0.747 & 0.631 & 0.790 & 0.863 & 79.199 & 0.907 \\
\textbf{\ODD}          & 0.096 & 0.796 & 0.679 & 0.828 & 0.968 & 83.147 & 0.928 \\
\midrule

\multicolumn{8}{c}{\textbf{\footnotesize Incremental Drift}} \\
\textbf{Greedy}        & 0     & 0.740 & 0.600 & 0.790 & 0.865 & 78.304 & 0.908 \\
\textbf{Temp Scaled}   & 0     & 0.725 & 0.574 & 0.775 & 0.862 & 76.886 & 0.902 \\
\textbf{\ODD}          & 0.052 & 0.803 & 0.682 & 0.838 & 0.976 & 83.826 & 0.935 \\
\midrule
\multicolumn{8}{c}{\textbf{\footnotesize Gradual Drift}} \\
\textbf{Greedy}        & 0     & 0.702 & 0.555 & 0.760 & 0.855 & 74.031 & 0.890 \\
\textbf{Temp Scaled}   & 0     & 0.687 & 0.529 & 0.745 & 0.853 & 72.710 & 0.885 \\
\textbf{\ODD}          & 0.037 & 0.790 & 0.663 & 0.825 & 0.971 & 82.824 & 0.928 \\
\bottomrule
\end{tabular}
\end{adjustbox}
\vspace{-4mm} 
\end{table}
\begin{figure}[ht!]
\centering
\begin{subfigure}[t]{\textwidth}
    \centering
    \includegraphics[width=0.95\textwidth, height=0.22\textheight]{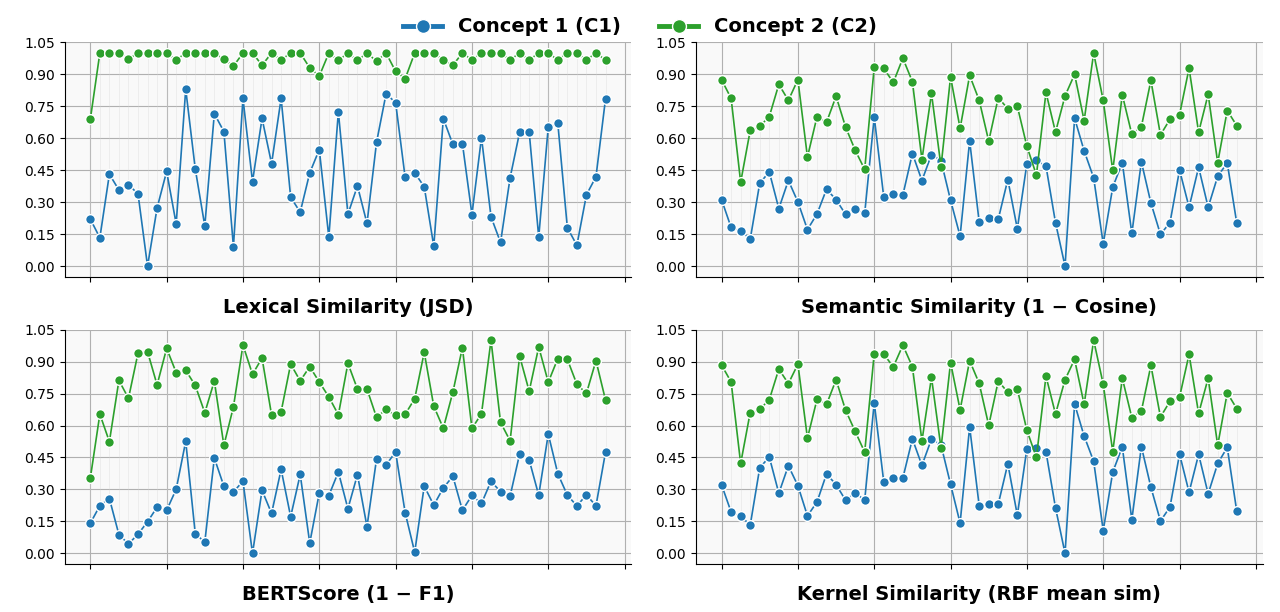}    
\end{subfigure}
\centerline{\rule{0.5\textwidth}{0.3pt}}
\begin{subfigure}[t]{\textwidth}
    \centering
    \includegraphics[width=0.95\textwidth, height=0.22\textheight]{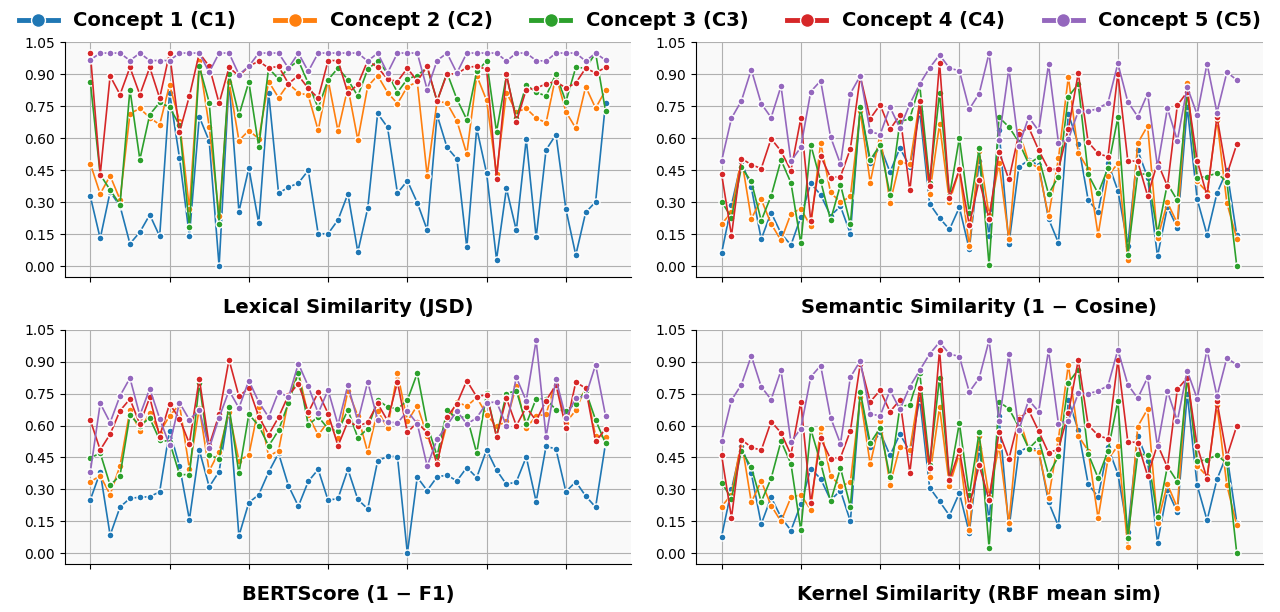}
\end{subfigure}
\centerline{\rule{0.5\textwidth}{0.3pt}}
\begin{subfigure}[t]{\textwidth}
    \centering
    \includegraphics[width=0.95\textwidth, height=0.22\textheight]{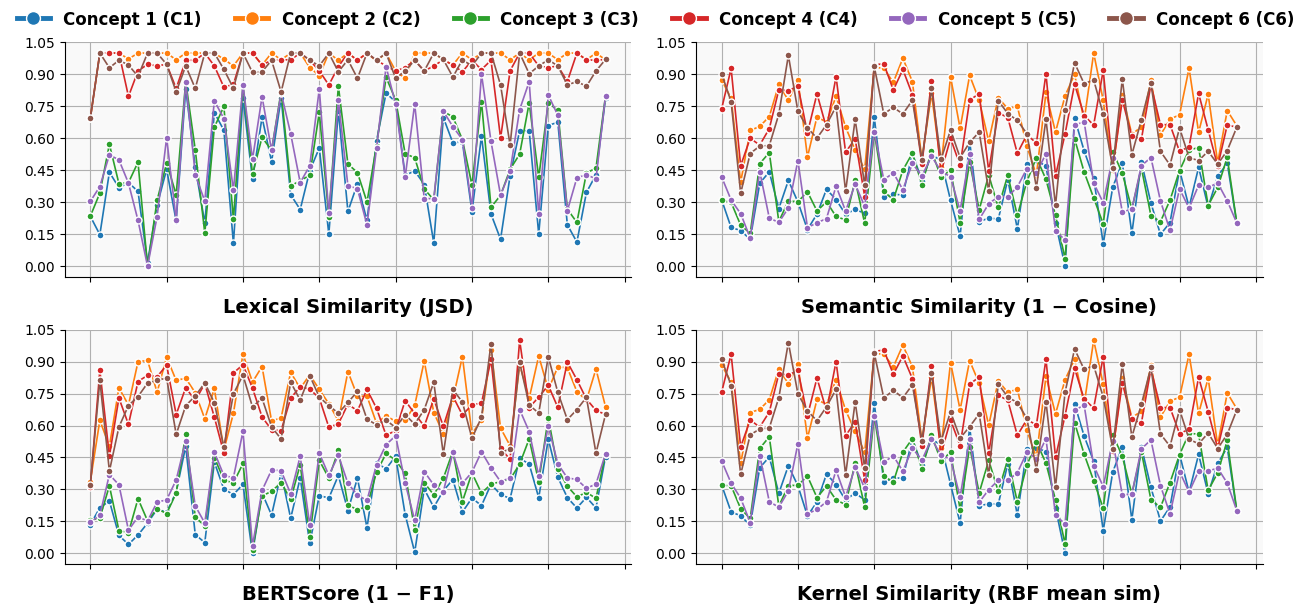}
\end{subfigure}
\vspace{-18pt}
\caption{\centering\scriptsize
Concept drift in placeholders: \textbf{(top)} Abrupt, \textbf{(middle)} Incremental, \textbf{(bottom)} Gradual.}
\label{fig:combined_drift_placeholders}
\vspace{-18pt}
\end{figure}

\subsection{Performance Analysis under Drift}
\vspace{-2pt}
Figure~\ref{fig:nlg_barplot_abrupt} reports mean \textbf{syntactic} and \textbf{semantic} performance under the \textit{abrupt} scenario across the decoding strategies: \textit{LLM-Greedy}, \textit{LLM-Temp Scaled}, and \ODD. The proposed \ODD consistently outperforms both baselines, with the largest gains in semantic alignment. It achieves a Cosine Similarity of \textbf{0.968} versus \textbf{0.865} and \textbf{0.863}, and a BERTScore of \textbf{0.928} versus \textbf{0.911} and \textbf{0.907} for \textit{LLM-Greedy} and \textit{LLM-Temp Scaled}, respectively. These results confirm \ODD’s robustness in preserving meaning and coherence under abrupt lexical shifts.

Table~\ref{tab:drift_results} reports mean \textbf{syntactic} and \textbf{semantic} performance for \textit{LLM-Greedy}, \textit{LLM-Temp Scaled}, and \ODD. Under \textit{incremental} drift, \ODD shows the strongest stability, achieving a Cosine Similarity of $\mathbf{0.976}$ versus $\mathbf{0.865}$ and $\mathbf{0.862}$, and a BERTScore of $\mathbf{0.935}$ versus $\mathbf{0.908}$ and $\mathbf{0.902}$. It also attains the best syntactic alignment (ChrF~$\mathbf{83.83}$ vs.~$\mathbf{78.30}$ and~$\mathbf{76.89}$). Under \textit{gradual} drift, \ODD again yields the highest semantic scores, with Cosine Similarity $\mathbf{0.971}$ (vs.~$\mathbf{0.855}$ and $\mathbf{0.853}$) and BERTScore $\mathbf{0.928}$ (vs.~$\mathbf{0.890}$ and $\mathbf{0.885}$). For \textit{syntactic} alignment, \ODD achieves the highest ChrF ($\mathbf{82.82}$ vs.\ $\mathbf{74.03}$ and $\mathbf{72.71}$), with BLEU and ROUGE-L showing the same trend. These results confirm \ODD’s resilience to gradual lexical drift while preserving both form and meaning.


Across all scenarios, \ODD consistently outperformed the baselines in \textit{syntactic} and \textit{semantic} quality. Under the most challenging \textit{gradual} drift, \ODD achieved a ROUGE-L of $0.825$ (an absolute gain of $\mathbf{0.065}$ over the strongest baseline) and a Cosine Similarity of $0.971$ (a relative improvement of $\mathbf{13.6\%}$). These gains, consistent across abrupt and incremental drift, demonstrate \ODD’s ability to maintain syntactic structure and semantic coherence as lexical distributions shift.

To illustrate the \textbf{qualitative} impact of \ODD, we highlight an example from the abrupt drift scenario. After the drift, the ground truth introduces a new plan (“Talk + Net Pack”) and brand (“TalkNow Crew”). Both baselines continue generating outdated Concept~1 templates such as “To sign up for a Mobile Voice and Data Communication Service...,” failing to reflect the new domain state. In contrast, \ODD{} adapts correctly and produces “To sign up for a Talk + Net Packet plan with TalkNow Crew...,” aligning with the drifted concept while maintaining structural consistency. This example shows how \ODD updates in real time while baseline decoders remain anchored to pre-drift patterns.
\subsection{Ablation Summary \& Reproducibility}
\vspace{-2pt}
\paragraph{Ablation Summary.}
Using both \textit{disagreement} ($\Omega_t$) and \textit{continuity} ($\Gamma_t$) jointly produced the most stable behavior; enabling only one signal reduced robustness. \textit{Temperature calibration} was also critical, as disabling it caused confidence-scale mismatches between the LLM and Trie      distributions. Trie feature weights were fixed to \textit{uniform values} ($\lambda_F=\lambda_L=\lambda_R=1/3$), which yielded stable performance; exploring alternative weightings remains future work.
\vspace{-11pt}
\paragraph{Reproducibility.}
All experiments used the \texttt{gpt2-medium} model (355M), fine-tuned for 10 epochs with a fixed \textit{60\% train, 10\% val, 30\% test} split. The Trie feature weights ($\lambda_F, \lambda_L, \lambda_R$) were set uniformly to $1/3$. The test sets for the abrupt, incremental, and gradual drift scenarios followed the concept progressions shown in Fig.~3, and all experiments were executed under identical configurations. Experiments ran on the Lonestar6 (LS6) system at TACC using a single NVIDIA A100 GPU (40GB), 64 CPU cores, and 64GB RAM.

\section{Conclusion}
\vspace{-5pt}
This work introduced \textbf{\ODD (Online Domain-Aware Decoding)}, pioneering the \textit{first inference-time, online domain-aware decoding framework} to mitigate concept drift without external retrieval or additional model training. \ODD is defined by a \textit{single-pass decoding architecture} powered by an \textit{Adaptive Fusion Mechanism}. This mechanism dynamically balances LLM fluency with trie constraints by applying a runtime reward–penalty bias based on \textit{disagreement} and \textit{continuity} signals, ensuring structural correctness and real-time adaptation with near-zero latency overhead. The framework’s efficiency relies on its \textit{lightweight prefix-trie} architecture. This trie stores tokenized placeholders, enabling fast $\mathcal{O}(L)$ updates, efficient $\mathcal{O}(L^2)$ retrieval, compact memory scaling, and near-zero latency via linear-time lookups during decoding. Empirically, using a unified \textit{Online Domain Shift (ODS)} benchmark covering abrupt, incremental, and gradual drift, \ODD demonstrated superior robustness across all scenarios. It outperformed baseline decoding methods on all NLG metrics, with an average absolute gain of $\mathbf{0.065}$ in ROUGE-L and a $\mathbf{13.6\%}$ relative gain in Cosine Similarity, underscoring its ability to adapt reliably to dynamic linguistic environments.
\vspace{-10pt}
\bibliographystyle{splncs04}
\bibliography{references}

@inproceedings{anderson2017guided,
  author    = {Peter Anderson and Basura Fernando and Mark Johnson and Stephen Gould},
  title     = {Guided Open Vocabulary Image Captioning with Constrained Beam Search},
  booktitle = {Proc.\ EMNLP ’17},
  year      = {2017},
  pages     = {936--945}
}

@article{brown2020language,
author = {Brown, T. and Mann, B. and Ryder, N. and Subbiah, M. and Kaplan, J. and et al.},
  title   = {Language Models are Few-Shot Learners},
  journal = {Adv.\ Neural Inf.\ Process.\ Syst.},
  volume  = {33},
  year    = {2020}
}

@article{chowdhery2023palm,
author = {Chowdhery, A. and Narang, S. and Devlin, J. and Bosma, M. and et al.},
  title   = {PaLM: Scaling Language Modeling with Pathways},
  journal = {J.\ Mach.\ Learn.\ Res.},
  volume  = {24},
  number  = {240},
  year    = {2023}
}

@article{touvron2023llama,
author = {Touvron, H. and Lavril, T. and Izacard, G. and Martinet, X. and et al.},
  title   = {LLaMA: Open and Efficient Foundation Language Models},
  journal = {arXiv:2302.13971},
  year    = {2023}
}

@article{ouyang2022training,
author = {Ouyang, L. and Wu, J. and Jiang, X. and Almeida, D. and et al.},
  title   = {Training Language Models to Follow Instructions with Human Feedback},
  journal = {Adv.\ Neural Inf.\ Process.\ Syst.},
  volume  = {35},
  year    = {2022}
}

@inproceedings{zhang2020dialogpt,
author = {Zhang, Y. and Sun, S. and Galley, M. and et al.},
  title     = {DIALOGPT: Large-Scale Generative Pre-training for Conversational Response Generation},
  booktitle = {ACL ’20 System Demonstrations},
  year      = {2020}
}

@article{ji2023survey,
author = {Ji, Z. and Lee, N. and Frieske, R. and Yu, T. and et al.},
  title   = {Survey of Hallucination in Natural Language Generation},
  journal = {ACM Comput.\ Surv.},
  volume  = {55},
  number  = {12},
  year    = {2023}
}

@inproceedings{maynez-etal-2020-faithfulness,
  author    = {Joshua Maynez and Shashi Narayan and Bernd Bohnet and Ryan McDonald},
  title     = {On Faithfulness and Factuality in Abstractive Summarization},
  booktitle = {Proc.\ ACL ’20},
  pages     = {1906--1919},
  year      = {2020}
}

@article{shi2024continual,
author = {Shi, H. and Xu, Z. and Wang, H. and Qin, W. and et al.},
  title   = {Continual Learning of Large Language Models: A Comprehensive Survey},
  journal = {ACM Comput.\ Surv.},
  year    = {2024}
}

@inproceedings{hulora,
  author    = {Edward J. Hu and Yelong Shen and Phillip Wallis and Zeyuan Allen-Zhu and Yuanzhi Li and Shean Wang and Lu Wang and Weizhu Chen},
  title     = {LoRA: Low-Rank Adaptation of Large Language Models},
  booktitle = {Proc.\ ICLR},
  year      = {2022}
}

@inproceedings{krause2021gedi,
author = {Krause, B. and Gotmare, A. and McCann, B. and et al.},
  title     = {GeDi: Generative Discriminator Guided Sequence Generation},
  booktitle = {Findings EMNLP ’21},
  pages     = {4929--4952},
  year      = {2021}
}

@article{gama2014survey,
  author  = {Jo{\~a}o Gama and Indr{\.e} {\v{Z}}liobait{\.e} and Albert Bifet and Mykola Pechenizkiy and Abdelhamid Bouchachia},
  title   = {A Survey on Concept Drift Adaptation},
  journal = {ACM Comput.\ Surv.},
  volume  = {46},
  number  = {4},
  year    = {2014}
}

@article{de2021continual,
author = {De Lange, M. and Aljundi, R. and Masana, M. and et al.},
  title   = {A Continual Learning Survey: Defying Forgetting in Classification Tasks},
  journal = {IEEE Trans.\ Pattern Anal.\ Mach.\ Intell.},
  year    = {2021}
}

@article{zhao2023survey,
  author  = {Wayne Xin Zhao and Kun Zhou and Junyi Li and Tianyi Tang and Xiaolei Wang and Yupeng Hou and Yingqian Min and Beichen Zhang and Junjie Zhang and Zican Dong and others},
  title   = {A Survey of Large Language Models},
  journal = {arXiv},
  year    = {2023}
}

@inproceedings{dathathriplug,
author = {Dathathri, S. and Madotto, A. and Lan, J. and et al.},
  title     = {Plug and Play Language Models: A Simple Approach to Controlled Text Generation},
  booktitle = {Proc.\ ICLR},
  year      = {2020}
}

@inproceedings{yang2021fudge,
  author    = {Kevin Yang and Dan Klein},
  title     = {FUDGE: Controlled Text Generation with Future Discriminators},
  booktitle = {Proc.\ NAACL-HLT ’21},
  pages     = {3511--3535},
  year      = {2021}
}

@inproceedings{liu2021dexperts,
  author    = {Liu, Alisa et al.},
  title     = {DExperts: Decoding-Time Controlled Text Generation with Experts and Anti-Experts},
  booktitle = {ACL-IJCNLP ’21},
  year      = {2021}
}

@article{schick2021self,
  author  = {Timo Schick and Sahana Udupa and Hinrich Sch{\"u}tze},
  title   = {Self-Diagnosis and Self-Debiasing: A Proposal for Reducing Corpus-Based Bias in NLP},
journal = {TACL},
  volume  = {9},
  pages   = {1408--1424},
  year    = {2021}
}

@article{l2016vocabulary,
  author  = {Gurvan L'Hostis and David Grangier and Michael Auli},
  title   = {Vocabulary Selection Strategies for Neural Machine Translation},
  journal = {arXiv:1610.00072},
  year    = {2016}
}

@inproceedings{wu2018neural,
  author    = {Yu Wu and Wei Wu and Dejian Yang and Can Xu and Zhoujun Li},
  title     = {Neural Response Generation with Dynamic Vocabularies},
  booktitle = {Proc.\ AAAI ’18},
  year      = {2018}
}

@inproceedings{liuknowledge,
  author    = {Ruibo Liu and Guoqing Zheng and Shashank Gupta and Radhika Gaonkar and Chongyang Gao and Soroush Vosoughi and Milad Shokouhi and Ahmed Hassan Awadallah},
  title     = {Knowledge Infused Decoding},
  booktitle = {Proc.\ ICLR},
  year      = {2023}
}

@inproceedings{li2021prefix,
  author    = {Xiang Lisa Li and Percy Liang},
  title     = {Prefix-Tuning: Optimizing Continuous Prompts for Generation},
  booktitle = {Proc.\ ACL-IJCNLP ’21},
  pages     = {4582--4597},
  year      = {2021}
}

@article{meng2022locating,
  author  = {Kevin Meng and David Bau and Alex Andonian and Yonatan Belinkov},
  title   = {Locating and Editing Factual Associations in {GPT}},
  journal = {Adv.\ Neural Inf.\ Process.\ Syst.},
  volume  = {35},
  pages   = {17359--17372},
  year    = {2022}
}

@inproceedings{meng2023mass,
  author    = {Kevin Meng and Arnab Sen Sharma and Alex J. Andonian and Yonatan Belinkov and David Bau},
  title     = {Mass-Editing Memory in a Transformer},
  booktitle = {Proc.\ ICLR},
  year      = {2023}
}

@inproceedings{khandelwalgeneralization,
  author    = {Urvashi Khandelwal and Omer Levy and Dan Jurafsky and Luke Zettlemoyer and Mike Lewis},
  title     = {Generalization through Memorization: Nearest Neighbor Language Models},
  booktitle = {Proc.\ ICLR},
  year      = {2020}
}

@article{lewis2020retrieval,
author = {Lewis, P. and Perez, E. and Piktus, A. and Petroni, F. and et al.},
  title   = {Retrieval-Augmented Generation for Knowledge-Intensive NLP Tasks},
  journal = {Adv.\ Neural Inf.\ Process.\ Syst.},
  volume  = {33},
  year    = {2020}
}

@inproceedings{borgeaud2022improving,
  author    = {Borgeaud, Sebastian et al.},
  title     = {Improving Language Models by Retrieving from Trillions of Tokens},  
  booktitle = {ICML ’22},
  pages     = {2206--2240},
  year      = {2022}
}

@inproceedings{lester2021power,
  author    = {Brian Lester and Rami Al-Rfou and Noah Constant},
  title     = {The Power of Scale for Parameter-Efficient Prompt Tuning},
  booktitle = {Proc.\ EMNLP ’21},
  pages     = {3045--3059},
  year      = {2021}
}

@misc{bitext_telco_llm,
  author = {{Bitext}},
  title  = {Bitext Telco LLM Chatbot Training Dataset},
  year   = {2023},
  note   = {Available on Hugging Face: \texttt{bitext/Bitext-telco-llm-chatbot-training-dataset}}
}

@inproceedings{rajpurkar2016squad,
  author    = {Pranav Rajpurkar and Jian Zhang and Konstantin Lopyrev and Percy Liang},
  title     = {SQuAD: 100{,}000+ Questions for Machine Comprehension of Text},
  booktitle = {Proc.\ EMNLP ’16},
  pages     = {2383--2392},
  year      = {2016}
}

@article{yujian2007normalized,
  author  = {Li Yujian and Liu Bo},
  title   = {A Normalized Levenshtein Distance Metric},
  journal = {IEEE Trans.\ Pattern Anal.\ Mach.\ Intell.},
  volume  = {29},
  number  = {6},
  pages   = {1091--1095},
  year    = {2007}
}

@inproceedings{papineni2002bleu,
  author    = {Kishore Papineni and Salim Roukos and Todd Ward and Wei-Jing Zhu},
  title     = {BLEU: A Method for Automatic Evaluation of Machine Translation},
  booktitle = {Proc.\ ACL ’02},
  pages     = {311--318},
  year      = {2002}
}

@inproceedings{lin2004rouge,
  author    = {Chin-Yew Lin},
  title     = {ROUGE: A Package for Automatic Evaluation of Summaries},
  booktitle = {Text Summarization Branches Out},
  pages     = {74--81},
  year      = {2004}
}

@inproceedings{reimers2019sentence,
  author    = {Nils Reimers and Iryna Gurevych},
  title     = {Sentence-{BERT}: Sentence Embeddings Using Siamese {BERT}-Networks},
  booktitle = {Proc.\ EMNLP-IJCNLP ’19},
  pages     = {3982--3992},
  year      = {2019}
}

@inproceedings{popovic2015chrf,
  author    = {Maja Popovi{\'c}},
  title     = {chrF: Character n-gram F-score for Automatic {MT} Evaluation},
  booktitle = {Proc.\ WMT ’15},
  pages     = {392--395},
  year      = {2015}
}

@inproceedings{zhang2020bertscore,
  author    = {Tianyi Zhang and Varsha Kishore and Felix Wu and Kilian Q. Weinberger and Yoav Artzi},
  title     = {BERTScore: Evaluating Text Generation with BERT},
  booktitle = {Proc.\ ICLR},
  year      = {2020}
}

@misc{ODD-GIT,
  author       = {Anonymous},
  title        = {ODD Online Domain-aware LLM Decoding for Continual Domain Evolution},
  year         = {2025},
  howpublished = {\url{https://github.com/anonymous273800/ODD}},
  note         = {GitHub repository}
}
\end{document}